\documentclass[runningheads]{llncs}

\usepackage{graphicx}
\usepackage{amssymb}
\usepackage{amsmath}
\usepackage{multirow}
\usepackage{float}
\usepackage{url}
\usepackage{array}

\begin{document}
\title{Satirical News Detection with Semantic Feature Extraction and Game-theoretic Rough Sets}
\titlerunning{Satirical News Detection with Semantic Feature Extraction and GTRS}

\author{Yue Zhou\inst{1} \and Yan Zhang\inst{1} \and JingTao Yao\inst{2}}
\authorrunning{Y. Zhou \&  Y. Zhang \& J. T. Yao}
\institute{School of Computer Science and Engineering\\
California State University, San Bernardino, CA  USA \and
Department of Computer Science, University of Regina, SK Canada\\
\email{Yue.Zhou@csusb.edu, Yan.Zhang@csusb.edu, jtyao@cs.uregina.ca}\\
}

\maketitle

\begin{abstract}

Satirical news detection is an important yet challenging task to prevent spread of misinformation. Many feature based and end-to-end neural nets based satirical news detection systems have been proposed and delivered promising results.
Existing approaches explore comprehensive word features from satirical news articles, but lack semantic metrics using word vectors for tweet form satirical news.
Moreover, the vagueness of satire and news parody determines that a news tweet can hardly be classified with a binary decision, that is, satirical or legitimate.
To address these issues, we collect satirical and legitimate news tweets, and propose a semantic feature based approach. Features are extracted by exploring inconsistencies in phrases, entities, and between main and relative clauses.
We apply game-theoretic rough set model to detect satirical news, in which probabilistic thresholds are derived by game equilibrium and repetition learning mechanism. Experimental results on the collected dataset show the robustness and improvement of the proposed approach compared with Pawlak rough set model and SVM.

\end{abstract}

\section{Introduction}
\label{sec:intro}

Satirical news, which uses parody characterized in a conventional news style, has now become an entertainment on social media. While news satire is claimed to be pure comedic and of amusement, it makes statements on real events often with the aim of attaining social criticism and influencing change~\cite{sterling2009encyclopedia}. Satirical news can also be misleading to readers, even though it is not designed for falsifications. Given such sophistication, satirical news detection is a necessary yet challenging natural language processing (NLP) task. Many feature based fake or satirical news detection systems \cite{burfoot2009automatic,rubin2016fake,shu2017fake} extract features from word relations given by statistics or lexical database, and other linguistic features. In addition, with the great success of deep learning in NLP in recent years, many end-to-end neural nets based detection systems~\cite{de2018attending,ruchansky2017csi,wang2018eann} have been proposed and delivered promising results on satirical news article detection.

However, with the evolution of fast-paced social media, satirical news has been condensed into a satirical-news-in-one-sentence form. For example, one single tweet of ``If earth continues to warm at current rate moon will be mostly underwater by 2400" by The Onion is largely consumed and spread by social media users than the corresponding full article posted on The Onion website. Existing detection systems trained on full document data might not be applicable to such form of satirical news. Therefore, we collect news tweets from satirical news sources such as The Onion, The New Yorker (Borowitz Report) and legitimate news sources such as Wall Street Journal and CNN Breaking News. We explore the syntactic tree of the sentence and extract inconsistencies between attributes and head noun in noun phrases. We also detect the existence of named entities and relations between named entities and noun phrases as well as contradictions between the main clause and corresponding prepositional phrase. For a satirical news, such inconsistencies often exist since satirical news usually combines irrelevant components so as to attain surprise and humor. The discrepancies are measured by cosine similarity between word components where words are represented by Glove~\cite{pennington2014glove}. Sentence structures are derived by Flair, a state-of-the-art NLP framework, which better captures part-of-speech and named entity structures~\cite{akbik2018coling}.

Due to the obscurity of satire genre and lacks of information given tweet form satirical news, there exists ambiguity in satirical news, which causes great difficulty to make a traditional binary decision.
That is, it is difficult to classify one news as satirical or legitimate with available information.
Three-way decisions, proposed by YY Yao, added an option - deferral decision in the traditional yes-and-no binary decisions and can be used to classify satirical news~\cite{yao2012outline,yao2016three}.
That is, one news may be classified as satirical, legitimate, and deferral.
We apply rough sets model, particularly the game-theoretic rough sets to classify news into three groups, i.e., satirical, legitimate, and deferral.
Game-theoretic rough set (GTRS) model, proposed by JT Yao and Herbert, is a recent promising model for decision making in the rough set context~\cite{yao2008game}.
GTRS determine three decision regions from a tradeoff perspective when multiple criteria are involved to evaluate the classification models~\cite{zhang2017multi}.
Games are formulated to obtain a tradeoff between involved criteria.
The balanced thresholds of three decision regions can be induced from the game equilibria.
GTRS have been applied in recommendation systems~\cite{azam2014game},  medical decision making~\cite{yao2015web}, uncertainty analysis~\cite{zhang2015determining}, and spam filtering~\cite{zhang2019three}.

We apply GTRS model on our preprocessed dataset and divide all news into satirical, legitimate, or deferral regions.
The probabilistic thresholds that determine three decision regions are obtained by formulating competitive games between accuracy and coverage and then finding Nash equilibrium of games.
We perform extensive experiments on the collected dataset, fine-tuning the model by different discretization methods and variation of equivalent classes.
The experimental result shows that the performance of the proposed model  is superior compared with Pawlak rough sets model and SVM.

\section{Related Work}

Satirical news detection is an important yet challenging NLP task. Many feature based models have been proposed. Burfoot et al. extracted features of headline, profanity, and slang using word relations given by statistical metrics and lexical database~\cite{burfoot2009automatic}. Rubin et al. proposed a SVM based model with five features (absurdity, humor, grammar, negative affect, and punctuation) for fake news document detection~\cite{rubin2016fake}. Yang et al. presented linguistic features such as psycholinguistic feature based on dictionary and writing stylistic feature from part-of-speech tags distribution frequency~\cite{yang2017satirical}. Shu et al. gave a survey in which a set of feature extraction methods is introduced for fake news on social media~\cite{shu2017fake}. Conroy et al. also uses social network behavior to detect fake news~\cite{conroy2015automatic}. For satirical sentence classification, Davidov et al. extract patterns using word frequency and punctuation features for tweet sentences and amazon comments~\cite{davidov2010semi}. The detection of a certain type of sarcasm which contracts positive sentiment with a negative situation by analyzing the sentence pattern with a bootstrapped learning was also discussed~\cite{riloff2013sarcasm}. Although word level statistical features are widely used, with advanced word representations and state-of-the-art part-of-speech tagging and named entity recognition model, we observe that semantic features are more important than word level statistical features to model performance. Thus, we decompose the syntactic tree and use word vectors to more precisely capture the semantic inconsistencies in different structural parts of a satirical news tweet.

Recently, with the success of deep learning in NLP, many researchers attempted to detect fake news with end-to-end neural nets based approaches. Ruchansky et al. proposed a hybrid deep neural model which processes both text and user information~\cite{ruchansky2017csi}, while Wang et al. proposed a neural network model that takes both text and image data~\cite{wang2018eann} for detection. Sarkar et al. presented a neural network with attention to both capture sentence level and document level satire~\cite{de2018attending}. Some research analyzed sarcasm from non-news text. Ghosh and Veale~\cite{ghosh2017magnets} used both the linguistic context and the psychological context information with a bi-directional LSTM to detect sarcasm in users' tweets. They also published a feedback-based dataset by collecting the responses from the tweets authors for future analysis.
While all these works detect fake news given full text or image content, or target on non-news tweets, we attempt bridge the gap and detect satirical news by analyzing news tweets which concisely summarize the content of news.

\section{Methodology}

In this section, we will describe the composition and preprocessing of our dataset and introduce our model in detail. We create our dataset by collecting legitimate and satirical news tweets from different news source accounts. Our model aims to detect whether the content of a news tweet is satirical or legitimate. We first extract the semantic features based on inconsistencies in different structural parts of the tweet sentences, and then use these features to train game-theoretic rough set decision model.

\subsection{Dataset}

We collected approximately 9,000 news tweets from satirical news sources such as The Onion and Borowitz Report and about 11,000 news tweets from legitimate new sources such as Wall Street Journal and CNN Breaking News over the past three years. Each tweet is a concise summary of a news article.
The duplicated and extreme short tweets are removed.A news tweet is labeled as satirical if it is written by satirical news sources and legitimate if it is from legitimate news sources. Table~\ref{tbl:instances} gives an example of tweet instances that comprise our dataset.

\begin{table}[!htbp]
\renewcommand{\arraystretch}{1.1}
\caption{Examples of instances comprising the news tweet dataset}
\label{tbl:instances}
\centering
\begin{tabular}{p{8cm}|l|l}
\hline
 Content  & Source & Label \\
 \hline
 The White House confirms that President Donald Trump sent a letter to North Korean leader Kim Jong Un. & CNN & 0\\
 \hline
Illinois Senate plans vote on bills that could become the state's first budget in more than two years. & WSJ & 0\\
 \hline
 Naked Andrew Yang emerges from time vortex to warn debate audience about looming threat Of automation.  & TheOnion & 1\\
 \hline
 New study shows majority of late afternoon sleepiness at Work caused by undetected carbon monoxide leak. & TheOnion & 1\\
 \hline
  Devin Nunes accuses witnesses of misleading American people with facts. & BorowitzReport & 1\\
 \hline
\end{tabular}
\end{table}

\subsection{Semantic Feature Extraction}

Satirical news is not based on or does not aim to state the fact. Rather, it uses parody or humor to make statement, criticisms, or just amusements. In order to achieve such effect, contradictions are greatly utilized. Therefore, inconsistencies significantly exist in different parts of a satirical news tweet. In addition, there is a lack of entity or inconsistency between entities in news satire. We extracted these features at semantic level from different sub-structures of the news tweet. Different structural parts of the sentence are derived by part-of-speech tagging and named entity recognition by Flair. The inconsistencies in different structures are measured by cosine similarity of word phrases where words are represented by Glove word vectors. We explored three different aspects of inconsistency and designed metrics for their measurements. A word level feature using tf-idf~\cite{salton1986introduction} is added for robustness.

\subsubsection{Inconsistency in Noun Phrase Structures }
One way for a news satire to obtain surprise or humor effect is to combine irrelevant or less jointly used attributes and the head noun which they modified. For example, noun phrase such as ``rampant accountability", ``posthumous apology", ``Vatican basement", ``self-imposed mental construct" and other rare combinations are widely used in satirical news, while individual words themselves are common. To measure such inconsistency, we first select all leaf noun phrases (NP) extracted from the semantic trees to avoid repeated calculation. Then for each noun phrase, each adjacent word pair is selected and represented by 100-dim Glove word vector denoted as $(v_{t},w_{t})$. We define the averaged cosine similarity of noun phrase word pairs as:
\begin{align}
\label{equ:snp}
S_{N\!P}=\frac{1}{T}\sum_{t=1}^{T}cos(v_{t},w_{t})
\end{align}
where $T$ is a total number of word pairs. We use $S_{N\!P}$ as a  feature to capture the overall inconsistency in noun phrase uses. $S_{N\!P}$ ranges from -1 to 1, where a smaller value indicates more significant inconsistency.

\subsubsection{Inconsistency Between Clauses}
Another commonly used rhetoric approach for news satire is to make contradiction between the main clause and its prepositional phrase or relative clause. For instance, in the tweet ``Trump boys counter Chinese currency manipulation $by$ adding extra zeros To \$20 Bills.", contradiction or surprise is gained by contrasting irrelevant statements provided by different parts of the sentence.
Let $q$ and $p$ denote two clauses separated by main/relative relation or preposition, and $(w_{1},w_{1},... w_{q})$ and $(v_{1},v_{1},... v_{p})$ be the vectorized words in $q$ and $p$. Then we define inconsistency between $q$ and $p$ as:
\begin{align}
\label{equ:sqp}
S_{Q\!P}=cos(\sum_{q=1}^{Q}w_{q},\sum_{p=1}^{P}v_{p}))
\end{align}
Similarly, the feature $S_{Q\!P}$ is measured by cosine similarity of linear summations of word vectors, where smaller value indicates more significant inconsistency.

\subsubsection{Inconsistency Between Named Entities and Noun Phrases}
Even though many satirical news tweets are made based on real persons or events, most of them lack specific entities. Rather, because the news is fabricated, news writers use the words such as ``man",``woman",``local man", ``area woman",``local family" as subject. However, the inconsistency between named entities and noun phrases often exists in a news satire if a named entity is included. For example, the named entity ``Andrew Yang" and the noun phrases ``time vortex" show great inconsistency than ``President Trump", "Senate Republicans", and ``White House" do in the legitimate news ``President Trump invites Senate Republicans to the White House to talk about the funding bill." We define such inconsistency as a categorical feature that:
\begin{align}
\label{equ:cnern}
C_{N\! E\! R\! N}=\begin{cases}
0 & \text{ if } S_{N\! E\! R\! N} < \bar{S}_{N\! E\! R\! N}\\
 1& \text{ if } S_{N\! E\! R\! N} \geq  \bar{S}_{N\! E\! R\! N} \\
-1 & \text{ if there's no named entity} \\
\end{cases}
\end{align}
 $S_{N\! E\! R\! N}$ is the cosine similarity of named entities and noun phrases of a certain sentence and $\bar{S}_{N\! E\! R\! N}$ is the mean value of $S_{N\! E\! R\! N}$ in corpus.
\subsubsection{Word Level Feature Using TF-IDF} We calculated the difference of tf-idf scores between legitimate news corpus and satirical news corpus for each single word. Then, the set $S_{voc}$ that includes most representative legitimate news words is created by selecting top 100 words given the tf-idf difference. For a news tweet and any word $w$ in the tweet, we define the binary feature $B_{voc}$ as:
\begin{align}
\label{equ:bvoc}
B_{voc}=\begin{cases}
1 & \text{ if } w\in S_{voc} \\
0&   \text{ otherwise} \\
\end{cases}
\end{align}

\subsection{GTRS Decision Model}
We construct a Game-theoretic Rough Sets model for classification given the extracted features. Suppose $E\subseteq U \times U$ is an equivalence relation on a finite nonempty universe of objects $U$, where $E$ is reflexive, symmetric, and transitive.
The equivalence class containing an object $x$ is given by $[x]=\{y\in U|xEy\}$.
The objects in one equivalence class all have the same attribute values.
In the satirical news context,
given an undefined concept $satire$, probabilistic rough sets divide all news into three pairwise disjoint groups i.e., the satirical group $POS(satire)$, legitimate group $NEG(satire)$, and deferral group $BND(satire)$, by using the conditional probability $Pr(satire|[x]) = \frac{|satire\cap[x]|}{|[x]|}$ as the evaluation function, and $(\alpha,\beta)$ as the acceptance and rejection thresholds~\cite{yao2011superiority,yao2012outline,yao2016three}, that is,
\begin{align}
\label{eq:bgprs}
    POS_{(\alpha,\beta)}(satire) & =\{x \in U \mid Pr(satire|[x]) \geq \alpha \},\nonumber \\
    NEG_{(\alpha,\beta)}(satire) & =\{x \in U \mid Pr(satire|[x]) \leq \beta \}, \nonumber \\
    BND_{(\alpha,\beta)}(satire) & =\{x \in U \mid \beta < Pr(satire|[x]) < \alpha \}.
\end{align}
Given an equivalence class $[x]$,
if the conditional probability $Pr(satire|[x])$ is greater than or equal to the specified acceptance threshold $\alpha$, i.e., $Pr(satire|[x])\geq \alpha$,
we accept the news in $[x]$ as $satirical$.
If $Pr(satire|[x])$ is less than or equal to the specified rejection threshold $\beta$, i.e., $Pr(satire|[x])\leq \beta$
we reject the news in $[x]$ as $satirical$, or we accept the news in $[x]$ as $legitimate$.
If $Pr(satire|[x])$ is between $\alpha$ and $\beta$, i.e.,  $\beta<Pr(satire|[x])<\alpha$,
we defer to make decisions on the news in $[x]$.
Pawlak rough sets can be viewed as a special case of probabilistic rough sets with $(\alpha,\beta)=(1,0)$.

Given a pair of probabilistic thresholds $(\alpha, \beta)$, we can obtain a news classifier according to Equation~(\ref{eq:bgprs}).
The three regions are a partition of the universe $U$,
\begin{align}
\label{equ:3way}
\pi_{(\alpha,\beta)}(Satire)=\{POS_{(\alpha,\beta)}(Satire), BND_{(\alpha,\beta)}(Satire),NEG_{(\alpha,\beta)}(Satire)\}
\end{align}
Then, the accuracy and coverage rate to evaluate the performance of the derived classifier are defined as follows~\cite{zhang2017multi},

\begin{align}
\label{eq:acc}
  Acc_{(\alpha, \beta)}(Satire)=
  \frac{|Satire \cap POS_{(\alpha,\beta)}(Satire)| + |Satire^c \cap NEG_{(\alpha, \beta)}(Satire)| }{|POS_{(\alpha, \beta)}(Satire)|+|NEG_{(\alpha, \beta)}(Satire)| }
 \end{align}

\begin{align}
\label{eq:cov}
Cov_{(\alpha, \beta)}(Satire)=
\frac{|POS_{(\alpha, \beta)}(Satire)|+|NEG_{(\alpha, \beta)}(Satire)|}{|U|}
\end{align}

The criterion coverage indicates the proportions of news that can be confidently classified. Next, we will obtain $(\alpha, \beta)$ by game formulation and repetition learning.
\subsubsection{Game Formulation}
We construct a game $G=\{O,S,u\}$ given the set of game players $O$, the set of strategy profile $S$, and the payoff functions $u$, where the accuracy and coverage are two players, respectively, i.e., $O=\{acc, cov\}$.

The set of strategy profiles $S=S_{acc}\times S_{cov}$, where $S_{acc}$ and $S_{cov} $ are sets of possible strategies or actions performed by players $acc$ and $cov$.
The initial thresholds are set as $(1,0)$. All these strategies are the changes made on the initial thresholds,
\begin{align}
\label{eq:strategies}
S_{acc}&=\{ \beta\mbox{ no change}, \beta\mbox{ increases }c_{acc}, \beta\mbox{ increases }2\times c_{acc}\},\nonumber\\
S_{cov}&=\{ \alpha\mbox{ no change}, \alpha\mbox{ decreases }c_{cov}, \alpha\mbox{ decreases }2\times c_{cov}\}.
\end{align}
$c_{acc}$ and $c_{cov}$ denote the change steps used by two players, and their values are determined by the concrete experiment date set.

\emph{Payoff functions.}
The payoffs of players are $u=(u_{acc},u_{cov})$, and $u_{acc}$ and $u_{cov}$ denote the payoff functions of players $acc$ and $cov$, respectively.
Given a strategy profile $p=(s, t)$ with player $acc$ performing $s$ and player $cov$ performing $t$, the payoffs of $acc$ and $cov$ are $u_{acc}(s, t)$ and $u_{cov}(s, t)$.
We use $u_{acc}(\alpha,\beta)$ and $u_{cov}(\alpha,\beta)$ to show this relationship.
The payoff functions $u_{acc}(\alpha,\beta)$ and $u_{cov}(\alpha,\beta)$ are defined as,
\begin{align}
\label{eq:pf}
u_{acc}(s,t)\Rightarrow u_{acc}(\alpha,\beta)&=Acc_{(\alpha, \beta)}(Satire),\nonumber\\
u_{cov}(s,t)\Rightarrow u_{cov}(\alpha,\beta)&=Cov_{(\alpha, \beta)}(Satire),
\end{align}
where $Acc_{(\alpha, \beta)}(Satire)$ and $Cov_{(\alpha, \beta)}(Satire)$ are the accuracy and coverage defined in Equations~(\ref{eq:acc}) and~(\ref{eq:cov}).

\emph{Payoff table.} We use payoff tables to represent the formulated game.
Table~\ref{tbl:payofftable} shows a payoff table example in which both players have 3 strategies defined in Equation~ref{eq:stategies}.
\begin{table*}[ht]
\footnotesize
\renewcommand{\arraystretch}{1.1}
\centering
\caption{An example of a payoff table}
\label{tbl:payofftable}

\begin{tabular}{|c|c| >{\centering\arraybackslash}m{1.05in} | >{\centering\arraybackslash}m{1.35in} |>{\centering\arraybackslash}m{1.41in} |l}
\cline{1-5}
\multicolumn{2}{|c|}{}& \multicolumn{3}{|c|}{$cov$} \\
\cline{3-5}
\multicolumn{2}{|c|}{}&  $\alpha$ & $\alpha \downarrow c_{cov}$& $\alpha \downarrow 2 c_{cov}$ \\ \cline{1-5}
\multicolumn{1}{|c|}{}&
\multicolumn{1}{|c|}{$\beta$} &
$\big{<}u_{acc}(\alpha, \beta),$ &
$\big{<}u_{acc}(\alpha - c_{cov}, \beta),$ & $\big{<}u_{acc}(\alpha - 2c_{cov}, \beta),$ & \\[3pt]
\multicolumn{1}{|c|}{} &
\multicolumn{1}{|c|}{} &
$u_{cov}(\alpha, \beta)\big{>}$ & $u_{cov}(\alpha - c_{cov}, \beta)\big{>}$ & $u_{cov}(\alpha - 2c_{cov}, \beta)\big{>}$ & \\[3pt]
\cline{2-5}
\multicolumn{1}{|c|}{$\mbox{ }acc\mbox{ }$}&
\multicolumn{1}{|c|}{$\beta \uparrow c_{acc}$} &
$\big{<}u_{acc}(\alpha, \beta + c_{acc}),$ & $\big{<}u_{acc}(\alpha - c_{cov}, \beta + c_{acc}),$ &
$\big{<}u_{acc}(\alpha- 2c_{cov}, \beta + c_{acc}),$ & \\[3pt]
\multicolumn{1}{|c|}{}&
\multicolumn{1}{|c|}{ }&
$u_{cov}(\alpha, \beta + c_{acc})\big{>}$ & $u_{cov}(\alpha-c_{cov}, \beta+c_{acc})\big{>}$ & $u_{cov}(\alpha-2c_{cov}, \beta + c_{acc})\big{>}$ &\\[3pt]
\cline{2-5}
\multicolumn{1}{|c|}{}&
\multicolumn{1}{|c|}{$\beta\uparrow 2c_{acc}$} &
$\big{<}u_{acc}(\alpha, \beta+2c_{acc}),$ & $\big{<}u_{acc}(\alpha-c_{cov}, \beta+2c_{acc}),$ &
$\big{<}u_{acc}(\alpha-2c_{cov}, \beta+2c_{acc}),$ & \\[3pt]
\multicolumn{1}{|c|}{}&
\multicolumn{1}{|c|}{ }&
$u_{cov}(\alpha, \beta+2c_{acc})\big{>}$ & $u_{cov}(\alpha-c_{cov}, \beta+2c_{acc})\big{>}$ & $u_{cov}(\alpha-2c_{cov}, \beta+2c_{acc})\big{>}$ &\\[3pt]
\cline{1-5}
\end{tabular}
\normalsize
\end{table*}
The arrow $\downarrow$ denotes decreasing a value and  $\uparrow$ denotes increasing a value.
On each cell, the threshold values are determined by two players.

\subsubsection{Repetition Learning Mechanism}
We repeat the game with the new thresholds until a balanced solution is reached. We first analyzes the pure strategy equilibrium of the game and then check if the stopping criteria are satisfied.

\emph{Game equilibrium.} The game solution of pure strategy Nash equilibrium is used to determine possible game outcomes in GTRS.
The strategy profile $(s_{i},t_{j})$ is a pure strategy Nash equilibrium, if
\begin{align}
\forall s^{'}_{i} \in S_{acc}, &u_{acc}(s_{i},t_{j}) \geqslant u_{acc}(s^{'}_{i},t_{j}),\mbox{where } s_{i} \in S_{acc} \wedge s^{'}_{i} \neq s_{i}, \nonumber\\
\forall t^{'}_{j} \in S_{cov}, &u_{cov}(s_{i}, t_{j}) \geqslant u_{cov}(s_{i},t^{'}_{j}), \mbox{where }
t_{j} \in S_{cov} \wedge t^{'}_{j} \neq t_{j}.
\end{align}
 This means that none of players would like to change his strategy or they would loss benefit if deriving from this strategy profile, provided this player has the knowledge of other player's strategy.

\emph{Repetition of games.} Assuming that we formulate a game, in which the initial thresholds are $(\alpha, \beta)$, and the equilibrium analysis shows that the thresholds corresponding to the equilibrium are $(\alpha^{*}, \beta^{*})$.
If the thresholds $(\alpha^{*}, \beta^{*})$ do not satisfy the stopping criterion, we will update the initial thresholds in the subsequent games.
The initial thresholds of the new game will be set as $(\alpha^{*}, \beta^{*})$.
If the thresholds $(\alpha^{*}, \beta^{*})$ satisfy the stopping criterion, we may stop the repetition of games.

\emph{Stopping criterion.} We define the stopping criteria so that the iterations of games can stop at a proper time.
In this research, we set the stopping criterion as within the range of thresholds, the increase of one player's payoff is less than the decrease of the other player's payoff.

\section{Experiments}


There are 8757 news records in our preprocessed data set.
We use Jenks natural breaks~\cite{jenks1967data} to discretize continuous variables $S_{N\!P}$ and $S_{Q\!P}$ both into five categories denoted by nominal values from 0 to 4, where larger values still fall into bins with larger nominal value. Let $D_{N\!P}$ and $D_{Q\!P}$ denote the discretized variables $S_{N\!P}$ and $S_{Q\!P}$, respectively. We derived the information table that only contains discrete features from our original dataset. A fraction of the information table is shown in Table~\ref{tbl:informationtable}.
\begin{table}[ht]
\caption{The Information Table }
\label{tbl:informationtable}
\centering
\begin{tabular}{c|c|c|c|c|c|l}
\hline
Id & $D_{N\!P}$ & $D_{Q\!P}$ & $C_{N\! E\! R\! N}$ & $B_{voc}$ & target \\
\hline
1 & 0 & 2 & 0 &0 &1\\
\hline
2 & 1 & 2 &0 &0 &1 \\
\hline
3 & 2 & 2 &0 &1 & 0 \\
\hline
4 & 2 & 4 &1 &1 & 0 \\
\hline
5 & 2 & 3 &0 &0& 1 \\
\hline
6 & 4 & 3 &-1 &1 & 0\\
\hline
7 &2 & 3 &0 &0 & 0 \\
\hline
8 & 3 & 2 &-1 &0 & 1 \\
\hline
\end{tabular}
\end{table}

The news whose condition attributes have the same values are classified in an equivalence class $X_i$. We derived 149 equivalence classes and calculated the corresponding probability $Pr(X_i)$ and condition
probability $Pr(Satire|X_i)$ for each $X_i$.
The probability $Pr(X_{i})$ denotes the ratio of the number of news contained in the equivalence class $X_i$ to the total number of news in the dataset, while the conditional probability $Pr(Satire|X_{i})$ is the proportion of news in $X_i$ that are satirical.
We combine the equivalence classes with the same conditional probability and reduce the number of equivalence classes to 108.
Table~\ref{tbl:summarydata} shows a part of the probabilistic data information about the concept {\it satire}.
\begin{table*}[hb]
\caption{Summary of the partial experimental data}
\label{tbl:summarydata}
\centering
 \begin{tabular}{*{11}{c}}
\hline
& $X_{1}$ & $X_{2}$ & $X_{3}$& $X_{4}$ & $X_{5}$ & $X_{6}$ & $X_{7}$ & $X_{8}$ & $X_{9}$ &......\\
\hline
$Pr(X_{i})$ & 0.0315 &0.0054 &0.0026 &0.0071 &0.0062 &0.0018 &0.0015 &0.0098 &0.0009 &......\\
$Pr(Satire|X_{i})$&  1&0.9787&	0.9565&	0.9516&	0.9444&	0.9375&	0.9231&	0.9186&	0.875&	 ......\\
\hline
&...... & $X_{100}$ & $X_{101}$ & $X_{102}$ &$ X_{103}$ & $X_{104}$ & $X_{105}$ & $X_{106}$ & $X_{107}$ & $X_{108}$\\
\hline
$Pr(X_{i})$   & ......&0.0121&0.0138&0.0095	&0.0065&0.0383&0.0078&0.0107&0.0163	&0.048\\
$Pr(Satire|X_{i})$ & ......&0.0283&0.0248&0.0241&0.0175&0.0149&0.0147&0.0106&0.007&0\\
\hline
\end{tabular}
\end{table*}

\subsection{Finding Thresholds with GTRS}
We formulated a competitive game between the criteria accuracy and coverage to obtain the balanced probabilistic thresholds with the initial thresholds $(\alpha, \beta)=(1,0)$ and learning rate 0.03. As shown in the payoff table Table \ref{tbl:pt},
\begin{table*}[htp]
\caption{The payoff table}
\label{tbl:pt}
\centering
\begin{tabular}{|cc|c|c|c|l}
\cline{1-5}
 & & \multicolumn{3}{|c|}{$cov$} \\
\cline{3-5}
& & $\alpha$ & $\alpha\downarrow 0.03$ & $\alpha\downarrow0.06$& \\ \cline{1-5}
\multicolumn{1}{|c|}{}&
\multicolumn{1}{|c|}{$\beta$} &
$<1,0.0795>$ &$<0.9986,0.0849>$ & $<0.9909,0.1008>$& \\
\cline{2-5}
\multicolumn{1}{|c|}{$acc$}&
\multicolumn{1}{|c|}{$\beta\uparrow 0.03 $} &
$<0.9868, 0.2337>$ & $<0.9866,0.2391>$ &$<0.9843,0.255>$& \\
\cline{2-5}
\multicolumn{1}{|c|}{}&
\multicolumn{1}{|c|}{$\beta\uparrow 0.06$ } &
$<0.9799,0.3130>$ &$<0.9799,0.3184>$& $<\textbf{0.9784,0.3343}>$& \\
\cline{1-5}
\end{tabular}
\end{table*}
the cell at the right bottom corner is the game equilibrium whose strategy profile is ($\beta$ increases 0.06, $\alpha$ decreases 0.06).
The payoffs of the players are (0.9784,0.3343).
We set the stopping criterion as  the increase of one player's payoff is less than the decrease of the other player's payoff when the thresholds are within the range.
When the thresholds change from (1,0) to (0.94, 0.06), the accuracy is decreased from 1 to 0.9784 but the coverage is increased from 0.0795 to 0.3343.
We repeat the game by setting $(0.94, 0.06)$ as the next initial thresholds.

The competitive games are repeated seven times.
The result is shown in Table~\ref{tbl:repetition}.
\begin{table*}[ht]
\renewcommand{\arraystretch}{1.1}
\caption{The repetition of game}
\label{tbl:repetition}
\centering
\begin{tabular}{c|c|c|c|c|c|c|l}
\hline
 & Initial$(\alpha, \beta)$ & Strategies & Result$(\alpha, \beta)$ & Payoffs & $\downarrow >\uparrow$\\
 \hline
 1 & (1, 0) & $( \beta\uparrow 0.03, \alpha\downarrow 0.03 )$ & (0.94, 0.06) &$<0.9784,0.3343>$ & $\times$\\
 \hline
 2& (0.94, 0.06)& $( \beta\uparrow 0.03, \alpha\downarrow 0.03)$ & (0.88, 0.12)& $<0.9586,0.4805>$& $\times$\\
 \hline
 3& (0.88, 0.12)& $( \beta\uparrow 0.03, \alpha\downarrow 0.03)$ & (0.82, 0.18)& $<0.9433,0.554>$& $\times$\\
 \hline
 4& (0.82, 0.18)& $( \beta\uparrow 0.03, \alpha\downarrow 0.03)$ & (0.76, 0.24)& $<0.9218,0.6409>$& $\times$\\
 \hline
  5& (0.76, 0.24)& $(\beta\uparrow 0.03, \alpha\downarrow 0.03)$ & (0.7, 0.3)& $<0.8960,0.7467>$& $\times$\\
 \hline
  6& (0.7, 0.3)& $(\beta\uparrow 0.03, \alpha\downarrow 0.03)$ & (0.64, 0.36)& $
  <0.8791,0.8059>$& $\times$\\
 \hline
   7& (0.64, 0.36)& $(\beta\uparrow 0.03, \alpha\downarrow 0.03)$ & (0.58, 0.42)& $<0.8524,0.8946>$& $\times$\\
    \hline
   8& (0.58, 0.42)& $(\beta\uparrow 0.03, \alpha\downarrow 0.03)$ & (0.52, 0.48)& $<0.8271,0.9749>$& $\times$\\
 \hline
\end{tabular}
\end{table*}
 After the eighth iteration, the repetition of game is stopped because the further changes on thresholds may cause the thresholds lay outside of the range $0 < \beta < \alpha <1$,
 and the final result is the equilibrium of the seventh game $(\alpha, \beta)=(0.52, 0.48)$.

\subsection{Results}
 We compare Pawlak rough sets, SVM, and our GTRS approach on the proposed dataset.
 Table~\ref{tbl:compare} shows the results on the experimental data.
 \begin{table}[!htbp]
\renewcommand{\arraystretch}{1.1}
\caption{Experimental results}
\label{tbl:compare}
\centering
\begin{tabular}{|c|c|c|c|c|l}
\hline
 & $(\alpha, \beta)$ & Accuracy & Coverage& Modified accuracy \\
  \hline
SVM & - & $78\%$ & $100\%$ & $78\%$\\
 \hline
 Pawlak & (1, 0) & $100\%$ & $7.95\%$ & $53.98\%$\\
 \hline
 GTRS & (0.58, 0.42) & $\textbf{82.71\%}$ & $\textbf{97.49\%}$& $81.89\%$\\
 \hline
\end{tabular}
\end{table}
 The SVM classifier achieved an accuracy of $78\%$ with a $100\%$ coverage.
 The Pawlak rough set model using $(\alpha, \beta)=(1,0)$ achieves a $100\%$ accuracy and a coverage ratio of $7.95\%$, which means it can only classify $7.95\%$ of the data. The classifier constructed by GTRS with $(\alpha, \beta)=(0.52, 0.48)$ reached an accuracy $82.71\%$ and a coverage $97.49\%$.
which indicates that $97.49\%$ of data are able to be classified with accuracy of $82.71\%$.
The remaining $2.51\%$ of data can not be classified without providing more information. To make our method comparable to other baselines such as SVM, we assume random guessing is made on the deferral region and present the modified accuracy. The modified accuracy for our approach is then $0.8271\times 0.9749 + 0.5 \times 0.0251 =81.89\%$. Our methods shows significant improvement as compared to Pawlak model and SVM.


\section{Conclusion}

In this paper, we propose a satirical news detection approach based on extracted semantic features and game-theoretic rough sets.
In our mode, the semantic features extraction captures the inconsistency in the different structural parts of the sentences and the GTRS classifier can process the incomplete information based on repetitive learning and the acceptance and rejection thresholds.
The experimental results on our created satirical and legitimate news tweets dataset show that our model significantly outperforms Pawlak rough set model and SVM.
In particular, we demonstrate our model's ability to interpret satirical news detection from a semantic and information trade-off perspective.
Other interesting extensions of our paper may be to use rough set models to extract the linguistic features at document level.

\bibliographystyle{splncs04}

\bibliography{refs}

\begin{thebibliography}{10}
\providecommand{\url}[1]{\texttt{#1}}
\providecommand{\urlprefix}{URL }
\providecommand{\doi}[1]{https://doi.org/#1}

\bibitem{akbik2018coling}
Akbik, A., Blythe, D., Vollgraf, R.: Contextual string embeddings for sequence
  labeling. In: Proceedings of the 27th International Conference on
  Computational Linguistics. pp. 1638--1649. Springer (2018)

\bibitem{azam2014game}
Azam, N., Yao, J.T.: Game-theoretic rough sets for recommender systems.
  Knowledge-Based Systems  \textbf{72},  96--107 (2014)

\bibitem{burfoot2009automatic}
Burfoot, C., Baldwin, T.: Automatic satire detection: Are you having a laugh?
  In: Proceedings of the 2009 International Conference on Natural Language
  Processing. pp. 161--164. ACL (2009)

\bibitem{conroy2015automatic}
Conroy, N.J., Rubin, V.L., Chen, Y.: Automatic deception detection: Methods for
  finding fake news. In: Proceedings of the Association for Information Science
  and Technology. pp.~1--4. Wiley Online Library (2015)

\bibitem{davidov2010semi}
Davidov, D., Tsur, O., Rappoport, A.: Semi-supervised recognition of sarcastic
  sentences in twitter and amazon. In: Proceedings of the 14th Conference on
  Computational Natural Language Learning. pp. 107--116. ACL (2010)

\bibitem{de2018attending}
De~Sarkar, S., Yang, F., Mukherjee, A.: Attending sentences to detect satirical
  fake news. In: Proceedings of the 27th International Conference on
  Computational Linguistics. pp. 3371--3380. Springer (2018)

\bibitem{ghosh2017magnets}
Ghosh, A., Veale, T.: Magnets for sarcasm: Making sarcasm detection timely,
  contextual and very personal. In: Proceedings of the 2017 Conference on
  Empirical Methods in Natural Language Processing. pp. 482--491 (2017)

\bibitem{jenks1967data}
Jenks, G.F.: The data model concept in statistical mapping. International
  Yearbook of Cartography  \textbf{7},  186--190 (1967)

\bibitem{pennington2014glove}
Pennington, J., Socher, R., Manning, C.D.: Glove: Global vectors for word
  representation. In: Empirical Methods in Natural Language Processing. pp.
  1532--1543. ACL (2014)

\bibitem{riloff2013sarcasm}
Riloff, E., Qadir, A., Surve, P., De~Silva, L., Gilbert, N., Huang, R.: Sarcasm
  as contrast between a positive sentiment and negative situation. In:
  Proceedings of the 2013 Conference on Empirical Methods in Natural Language
  Processing. pp. 704--714 (2013)

\bibitem{rubin2016fake}
Rubin, V., Conroy, N., Chen, Y., Cornwell, S.: Fake news or truth? using
  satirical cues to detect potentially misleading news. In: Proceedings of the
  2nd Workshop on Computational Approaches to Deception Detection. pp. 7--17.
  ACM (2016)

\bibitem{ruchansky2017csi}
Ruchansky, N., Seo, S., Liu, Y.: Csi: A hybrid deep model for fake news
  detection. In: Proceedings of the 2017 Conference on Information and
  Knowledge Management. pp. 797--806. ACM (2017)

\bibitem{salton1986introduction}
Salton, G., McGill, M.J.: Introduction to Modern Information Retrieval.
  McGraw-Hill (1986)

\bibitem{shu2017fake}
Shu, K., Sliva, A., Wang, S., Tang, J., Liu, H.: Fake news detection on social
  media: A data mining perspective. ACM SIGKDD Explorations Newsletter
  \textbf{19}(1),  22--36 (2017)

\bibitem{sterling2009encyclopedia}
Sterling, C.H.: Encyclopedia of Journalism. Sage Publications (2009)

\bibitem{wang2018eann}
Wang, Y., Ma, F., Jin, Z., Yuan, Y., Xun, G., Jha, K., Su, L., Gao, J.: Eann:
  Event adversarial neural networks for multi-modal fake news detection. In:
  Proceedings of the 24th SIGKDD International Conference on Knowledge
  Discovery \& Data Mining. pp. 849--857. ACM (2018)

\bibitem{yang2017satirical}
Yang, F., Mukherjee, A., Dragut, E.: Satirical news detection and analysis
  using attention mechanism and linguistic features. arXiv preprint
  arXiv:1709.01189  (2017)

\bibitem{yao2008game}
Yao, J.T., Herbert, J.P.: A game-theoretic perspective on rough set analysis.
  Journal of Chongqing University of Posts and Telecommunications
  \textbf{20}(3),  291--298 (2008)

\bibitem{yao2015web}
Yao, J.T., Azam, N.: Web-based medical decision support systems for three-way
  medical decision making with game-theoretic rough sets. IEEE Transactions on
  Fuzzy Systems  \textbf{23}(1),  3--15 (2015)

\bibitem{yao2011superiority}
Yao, Y.Y.: The superiority of three-way decisions in probabilistic rough set
  models. Information Sciences  \textbf{181}(6),  1080--1096 (2011)

\bibitem{yao2012outline}
Yao, Y.Y.: An outline of a theory of three-way decisions. In: Proceedings of
  International Conference on Rough Sets and Current Trends in Computing. pp.
  1--17. Springer (2012)

\bibitem{yao2016three}
Yao, Y.Y.: Three-way decisions and cognitive computing. Cognitive Computation
  \textbf{8}(4),  543--554 (2016)

\bibitem{zhang2019three}
Zhang, Y., Liu, P.F., Yao, J.T.: Three-way email spam filtering with
  game-theoretic rough sets. In: Proceedings of the 2019 International
  Conference on Computing, Networking and Communications. pp. 552--556. IEEE
  (2019)

\bibitem{zhang2015determining}
Zhang, Y., Yao, J.T.: Determining three-way decision regions by combining gini
  objective functions and gtrs. In: Rough Sets, Fuzzy Sets, Data Mining, and
  Granular Computing. pp. 414--425. Springer (2015)

\bibitem{zhang2017multi}
Zhang, Y., Yao, J.T.: Multi-criteria based three-way classifications with
  game-theoretic rough sets. In: Proceedings of International Symposium on
  Methodologies for Intelligent Systems. pp. 550--559. Springer (2017)

\end{thebibliography}

\end{document}